# ATCSpeech: a multilingual pilot-controller speech corpus from real Air Traffic Control environment

*Bo Yang[1], Xianlong Tan[2], Zhengmao Chen[1], Bing Wang[2], Dan Li[2], Zhongping Yang[3], Xiping Wu[1], Yi Lin[1]\**

[1]National Key Laboratory of Air Traffic Control Automation System Technology, Sichuan University, Chengdu, 610065, China
[2] Southwest Air Traffic Management Bureau, Civil Aviation Administration of China (CAAC), Chengdu, 610000, China
[3] Wisesoft Co., Ltd., Chengdu, 610045, China

```
boyang@scu.edu.cn, caactxl@sina.com, chenzhengmao@scu.edu.cn，yzpping@126.com,
   kele4808@163.com, 13880237341@126.com, wuxipingstar@126.com, yilin@scu.edu.cn
                           * Corresponding author
```

## Abstract

Automatic Speech Recognition (ASR) is greatly developed in recent years, which expedites many applications on other fields. For the ASR research, speech corpus is always an essential foundation, especially for the vertical industry, such as Air Traffic Control (ATC). There are some speech corpora for common applications, public or paid. However, for the ATC, it is difficult to collect raw speeches from real systems due to safety issues. More importantly, for a supervised learning task like ASR, annotating the transcription is a more laborious work, which hugely restricts the prospect of ASR application. In this paper, a multilingual speech corpus (ATCSpeech) from real ATC systems, including accented Mandarin Chinese and English, is built and released to encourage the non-commercial ASR research in ATC domain. The corpus is detailly introduced from the perspective of data amount, speaker gender and role, speech quality and other attributions. In addition, the performance of our baseline ASR models is also reported. A community edition for our speech database can be applied and used under a special contrast. To our best knowledge, this is the first work that aims at building a real and multilingual ASR corpus for the air traffic related research.

**Index Terms**: automatic speech recognition, air traffic control, multilingual, ATCSpeech, speech corpus,

## 1. Introduction

Automatic speech recognition (ASR) is always a useful interface for human-machine interaction, which also becomes a promising technique for air traffic control (ATC). The ASR can be applied in the following ATC related scenes:

a) Real-time controlling speeches can be translated by an ASR system to detect a wealth of situational context information ( such as controlling intent), based on which the repetition can be automatically confirmed to relieve the controllers' workload and improve the traffic operational safety [1].

b) A robotic pilot can be implemented by combining the ASR with text-to-speech (TTS), which greatly reduces the cost of training air traffic controllers [2].

c) The ASR plays an important role in analyzing the historical ATC speech, such as traffic operation evaluation and reassignment, searching events from voice record devices [3].

Currently, almost all state-of-the-art ASR models are deep learning ones, in which the distribution with respect to the speech frame and text label is fitted by large-scale labelled dataset directly. Consequently, the ASR performance heavily depends on the dataset. Due to the air transportation safety and intellectual property issues, the real ATC speech is hard to collect. In addition, the domain specificities make it time-consuming and expensive to label sufficient samples for training a practical ASR system for ATC applications.

Compared to the common speech corpus, the ATC speech provides the following challenges for the ASR task, which should be considered in the research.

a) Volatile background noise and inferior intelligibility: a controller usually communicates with several pilots through a same radio frequency. Therefore, the noise model of the ATC speech is changing as the speaker changes. Moreover, the radio transmission is always a trouble for collecting high quality speeches.

b) Unstable speech rate: In general, the speech rate of ATC speech is higher than that of in daily life. However, it is affected by real-time working conditions and shows huge difference.

c) Multilingual ASR: In general, English is the universal language in ATC communication, whereas domestic pilots speak with controllers in local languages. Thus, the practical ATC speech may be multilingual.

d) Code switching: Code-switching is applied to eliminate misunderstanding by homonyms or near-homonyms in the communication, such as "零->洞" or "nine->niner".

e) Vocabulary imbalance: In practice, some out-of-vocabulary (OOV) words are existing in the ATC speeches since speakers do not comply with the communication procedures strictly. Moreover, the frequency of some special waypoint is also lower in the corpus. This leads to a serious situation that the frequency of different words in the corpus is extremely unbalanced, i.e., the sample sparsity.

In this paper, we strive to create a dedicated corpus for training practical ASR systems in ATC domain, in which the mentioned specificities are reflected to improve the overall quality. Several baseline ASR systems are also reported in this work. Our project can be traced back to October 2016, when a runway incursion occurred in the Shanghai Hongqiao international airport, China. The post-event analysis reported that the incident was expected to be detected if the real-time pilot-controller speech can be correctly translated as the input of a safety monitoring system. From then on, we are deeply aware of the importance of ASR in ATC and its promising application prospect. As a foundation this project, we established a team of 40 people to collect and label the real-time ATC speech for the ASR research.

In order to advocate the 'free data' movement and make contributions to the research community in ATC, we plan to share our corpus with no-commercial researchers and institutes. As far as we know, this is the first work that aims at creating a real ASR corpus for the ATC application with accented Chinese and English Speeches. To protect air transportation safety, a community edition of our corpus (about 40-hours Chinese speech and 19-hours English speech) is opened for publicly available at this time. The access permit for the released corpus can be freely applied and must be used in accordance with a special contract strictly. If someone wants to apply a data access permit, please contact our service department. Additional data service or the application of our full corpus can also be provided by other terms and contracts. The baseline ASR models with some test samples are also published.

The rest of this paper is organized as follows. The speech corpora related to common application and ATC are reviewed in Section 2. Section 3 provides the detail features of our corpus. The ASR performance of several baseline models are reported in Section 4. A short summary is in Section 5.

## 2. Existing ASR Corpora

It is generally known that the ASR performance highly depends on the training corpus due to its intrinsic supervised learning essence. Researchers all over the world have been striving to build available training corpus all the time. Several ASR corpora for common and ATC applications were found in the literature. Although some of them are publicly available, massive training samples are still difficult to obtain due to the complexity of the speech signal.

For the common ASR application, the CSLT at Tsinghua University, China released a 30-hours Chinese corpus, named (THCHS-30) [4]. The raw speeches were generated when speakers were reading newspapers at a silent office. The AISHELL also published two Mandarin Chinese corpora, named V1 [5] and V2 [6], which strives to transform the Chinese ASR task into an industrial scale. The LibriSpeech [7] published a large-scale English corpus (about 1000-hours speeches generated by reading novels), which contains clean and noisy subsets. The corpus comprises three scales: 100, 360 and 500 hours. The TED-LIUM [8] is also a popular corpus, which records the English-language TED talks. It has been updated to the third release for improving the data amount and speaker adaptation. Other ASR corpus can be found at [9], along with a ASR implementation under Kaldi framework [10].

For ATC applications, Delpech et al. reviewed several military ATC speech corpus in [11], including HIWIRE [12], nnMTAC [13]. Other datasets, VOCALISE [14] and air-ground communication [15], are also ATC related speech corpus. The ATCOSIM [16] simulated the controlling speech without the radio transmission noise. It contains about 10.7 hours data and is publicly available for all researchers. The LDC94S14A [17] is a real ATC speech collected from three US airports, about 70 hours, however, it was generated in the 1990s. Recently, an ASR challenge on accented English speech was held by Airbus, and 22 teams reported their results on given dataset [18]. Although there are several ATC related corpora, most of them are monolingual, simulated and very old data or need to be purchased at high price. Therefore, building dedicated corpus with accented multilingual speech and ATC-related elements is very important to the ASR application in ATC, this is what we strive to do in this work.

## 3. Data Features

### 3.1. Summary

Almost all the speeches in this corpus are collected from the voice record devices of real ATC systems in China. Therefore, the raw speeches are spoken in Chinese and English, and with real radio transmission noise. In our corpus, a very small part of the speeches are downloaded from [19], where the raw speeches are published without any transcription. In short, our database is a multilingual industrial ASR corpus in ATC domain.

The raw data is monaural speech with 8000 Hz sample rate and 16 bits sample size. Each training sample has a file pair: wave and label (transcription), which correspond to the input and output of an ASR model, respectively. There are about 58 hours speech data in our released corpus, which can be freely applied to a non-commercial research under a special contrast. Moreover, the common MFCC features of the raw speech signal are published for the training and dev dataset instead of the wave files, while the raw wave files are published for test dataset.

The transcription of the corpus is manual labelled, which is a human readable sentence (Chinese character and English word). There are about 698 Chinese characters and 584 English words in the corpus. In addition, to show the ASR specificities in ATC domain, some other attributions of the raw speeches are also published with the transcriptions, which are summarized as below:

a) Speaker gender: male (M) and female (F).
b) Speaker role: pilot (P) and controller (C).
c) Speech quality: clean (C) and noise (N).
d) Flight phase: ground (GND), tower (TWR), approach (APP) and en-route area control center (ACC).
e) Areas: indicting which airport the data is collected from. In this work, the data areas are encoded as the index.

Table 1: *Data size of the corpus.*

| Language | Train | | Dev | | Test | |
|---|---|---|---|---|---|---|
| | #U | #H | #U | #H | #U | #H |
| Chinese | 43186 | 37.77 | 1200 | 1.04 | 1200 | 1.03 |

| English | 15282 | 16.84 | 850 | 0.95 | 807 | 0.89 |

The dataset comprises of three subsets: training, dev and test, as shown in Tab. 1. The #H and #U denote the total duration (hours) and utterance for certain datasets, respectively. The data samples of each part are randomly selected from the available samples, in which we mainly focus on the diversity of the area, speaker gender, role and speech quality. Based on our data organization, at least one sample should be selected from each folder for the dev and test set. It should be noted that the data selection procedure for dev and test dataset is executed when receiving a new batch data from data team, not for the whole samples.

### 3.2. Speech

In the light of the ASR challenges in ATC domain, we report the statistics of our released ATCSpeech dataset, concerning the data size, utterance, flight phase, area, speaker role, gender, speech quality and speech rate. The detail information is summarized in the Tab. 2. The speech quality is a subjective evaluation based on the human hearing.

Table 2: *Statistics of ATCSpeech corpus.*

| Items | Chinese | English | Total |
|---|---|---|---|
| Amount | | | |
| #Number (Hours) | 39.83 | 18.69 | 58.52 |
| #Utterance | 45586 | 16939 | 62525 |
| Flight phase (Hours) | | | |
| TWR | 36.48 | 17.5 | 53.98 |
| APP | 3.47 | 1.19 | 4.66 |
| Area (Hours) | | | |
| TWR1 | 11.38 | 5.00 | 16.38 |
| TWR2 | 13.77 | 5.00 | 18.77 |
| TWR3 | 11.29 | 7.50 | 18.79 |
| APP1 | 3.45 | 1.19 | 4.64 |
| Speaker role (Hours) | | | |
| Polit | 21.12 | 8.92 | 30.04 |
| Controller | 18.73 | 9.77 | 28.50 |
| Speaker gender (Hours) | | | |
| Male | 36.16 | 16.94 | 53.10 |
| Female | 3.69 | 1.75 | 5.44 |
| Speech quality (Hours) | | | |
| Clean | 39.79 | 18.68 | 58.47 |
| Noisy | 0.05 | 0.01 | 0.06 |
| Others | | | |
| MoD (s) | 3.15 | 3.97 | - |
| SoD (s) | 1.04 | 1.36 | - |
| MoR (word/s) | 5.21 | 3.33 | - |
| SoR(word/s) | 1.22 | 0.72 | - |

MoD: mean of the duration. SoD: standard deviation of the duration. MoR: mean of the speech rate. SoR: standard deviation of the speech rate.

It can be seen from the result that the released corpus contains the raw speeches collected from different flight phase and controlling center, i.e., three towers and one approach center. The number of female speakers is obviously less than that of the male speakers due to the special duties of the air traffic controller. The data amount of the pilot's speech is slightly more than that of the controller. Due to the limitation of human hearing, only a little noisy data is labelled. In the released corpus, there are 5864 and 2330 flights in the Chinese and English speeches, respectively. The varieties of flight call-code indicates that our released corpus covers sufficient ATC-related information.

Analyzing the duration and rate of the speeches, the special challenges of ASR in ATC domain can also be confirmed, i.e., basically high speech rate with huge difference. As a comparison, the MoR and SoR of the open Chinese corpus THCHS30 are 3.48, 0.47 respectively, while the measurements for the open English corpus Librispeech are 2.73, 0.47, respectively.

### 3.3. Vocabulary

In this version, there are 698 Chinese characters and 584 English words in our word vocabulary. There are several special English words are in our corpus, such as the waypoint name PIKAS, P127. Moreover, there are about 60 English words in Chinese speeches, such as alpha, zulu. Similarly, some Chinese greeting words are also in the English speeches, such as nihao, xiexie. In general, the ASR is a sequential classification task, in which the probability of each frame belongs to a certain label class is predicted. Therefore, the class balance in the vocabulary is important to train a robust ASR model. We report the occurrence frequency of the lexicon in the Fig. 1 and Fig.2 for the Chinese and English speech, respectively. It can be seen that almost half of the words appear less than 10 times, i.e., the label classes are extremely unbalanced in this corpus. As illustrated before, the unbalanced vocabulary is also a key issue for the ASR task in ATC.

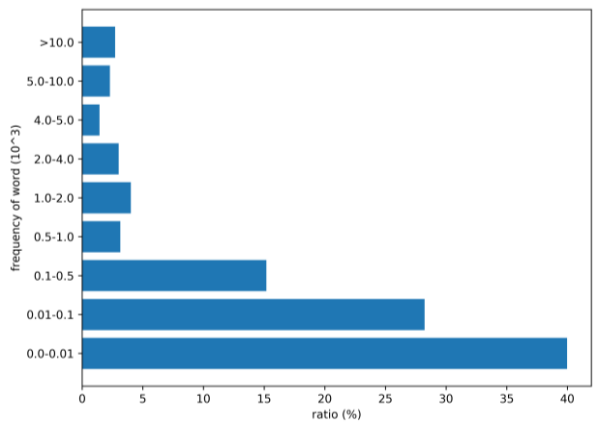

Figure 1: *Word frequency of Chinese speech.*

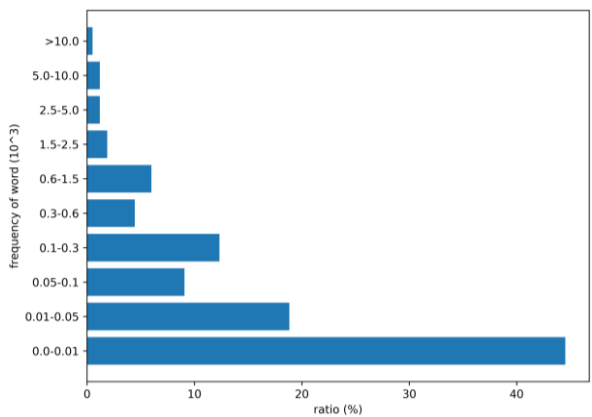

Figure 2: *Word frequency of English speech.*

# 4. Baseline ASR Systems

In this section, the experimental results of our baseline ASR systems are reported, which are built with the released ATCSpeech corpus. Based on the baseline systems, researchers can improve the model performance to address the ASR specificities in ATC domain, which is what we're trying to do to make our data publicly available.

## 4.1. Experiment configurations

Due to the widespread applications of the deep learning model, the baseline approaches in this work are also deep learning ones. Deep learning models learn the data distributions by huge training samples and have showed state-of-the-art performance on common ASR task. All the three approaches are the Connectionist Temporal Classification (CTC) based models. The three baseline models are implemented by referring Deep speech 2 (DS2) [20], Japser [21], and Wav2letter++ [22], respectively. The training, dev and test dataset are same as that of in Tab. 1. The All baseline models are trained separately for the Chinese and English speech. The mentioned baseline models are optimized by the same data division to ensure the experimental fairness. The output vocabulary of the baseline models for Chinese speech is Chinese character and English letter, while it is English letter for English speech. A N-best (10) decoding strategy is applied to correct the spelling errors based on a N-gram character language model (LM), in which the beam width is set to 20. The LM is trained by the transcriptions in the training dataset, and all baselines model use a same LM. The order of the LM are 9 and 18 for Chinese and English, respectively.

In this work, deep learning models are constructed based on the open framework Keras with backend of TensorFlow. The training server is configured as follows: 2*Intel Core i7-6800K, 2*NVIDIA GeForce GTX 1080Ti and 64 GB memory with operation system Ubuntu 16.04. The Adam optimizer with default implementation is used to train the three baselines. The training parameters for each baseline models are listed in Tab. 3.

Table 3: *Parameters for model training.*

| Baseline | language | batch size | #parameter (M) | #out labels |
|---|---|---|---|---|
| DS2 | Chinese | 64 | 26 | 679 |
|  | English |  |  | 28 |
| Jasper 10*3 | Chinese | 32 | 199 | 679 |
|  | English |  |  | 28 |
| Wav2letter++ | Chinese | 64 | 107 | 679 |
|  | English |  |  | 28 |

## 4.2. Baseline performance

Under certain experiment configurations, we train the baseline models which are then evaluated by a same test dataset. The final performance is measured by the character error rate (CER %) based on the Chinese character and English letter. Both the greedy decoding (AM) and beam search decoding (AM+LM) for each method are evaluated, in which experimental results for Chinese and English speech are reported in Tab. 4 and 5, respectively.

Table 4: *Baseline performance on Chinese speech.*

| Baselines | Training | | Test | |
|---|---|---|---|---|
|  | Loss | Epoch | AM | AM+LM |
| DS2 | 0.53 | 33 | 8.1 | 6.3 |
| Jasper 10*3 | 2.45 | 101 | 11.3 | 9.6 |
| Wav2letter++ | 2.37 | 136 | 14.3 | 12.5 |

Table 5: *Baseline performance on English speech.*

| Baselines | Training | | Test | |
|---|---|---|---|---|
|  | Loss | Epoch | AM | AM+LM |
| DS2 | 0.54 | 107 | 10.4 | 9.2 |
| Jasper 10*3 | 0.91 | 200 | 9.3 | 8.1 |
| Wav2letter++ | 1.06 | 307 | 11.3 | 10.1 |

From the experimental results we can see that all the popular ASR models are worked in our released corpus, for both the Chinese and English speech. The DS2 and Jasper based model obtain better performance for translating the Chinese and English speech, respectively. In addition, different baselines need different training epochs to obtain the convergence depending on their model architectures. Basically, because of the RNN architecture, the DS2 based baselines needs more time to train a same data iteration, but it can be converged with less iterations. On the contrary, the training time of other two baselines are less than that of the DS2 due to its CNN architecture. However, they need more training iterations to obtain the model convergence, as shown as the 'training epoch' in Tab.4 and 5.

# 5. Conclusions

In this paper, a real, multilingual ATC related ASR corpus, called as ATCSpeech, is released to promote the ASR research in this filed. The raw speeches spoken in accented Mandarin Chinese and English are collected from real air traffic control systems, which covers the ASR specificities to improve the overall quality of the corpus. The corpus can be applied and used in non-commercial researches. Additional attributions, flight phase, speaker gender and role, and speech quality, are also summarized in the transcription. The building details and CER based accuracy of baseline ASR systems are reported in this paper. Experimental results show that the state-of-the-art ASR models are worded on this released corpus. Based on this corpus, the ASR technique can be further used to make better decisions and continuous performance improvements in the aviation safety domain.

To the best of our knowledge, this is the first corpus that can be applied to train a practical ATC related ASR system. We sincerely hope the released corpus will benefit the new researchers and promote a promising prospect of ASR application in ATC domain. In addition, we also hope that more ASR systems can be proposed and improved based on this dataset, which further invokes more innovation and collaboration in the research community.

# 6. Acknowledgements

This work was jointly supported by the National Science Foundation of China (NSFC) and Civil Aviation Administration of China (CAAC) under the project No. U1833115. The authors would like to thank all contributors of the open ASR resources. Finally, the authors sincerely thank all the members of the sample producing team. It is their hard work that benefits the ASR research in the ATC domain.